# A VIKOR AND TOPSIS FOCUSED REANALYSIS OF THE MADM METHODS BASED ON LOGARITHMIC NORMALIZATION

**Sarfaraz Hashemkhani Zolfani[1], Morteza Yazdani[2], Dragan Pamucar[3], Pascale Zarate[4]**

[1]School of Engineering, Catholic University of the North, Larrondo, Coquimbo, Chile
[2]Department of Management, Universidad Loyola Andalucia, Seville, Spain
[3]Department of logistics, Military academy, University of Defence, Belgrade, Serbia
[4]University of Toulouse, IRIT, Toulouse, France

**Abstract**. *Decision and policy-makers in multi-criteria decision-making analysis take into account some strategies in order to analyze outcomes and to finally make an effective and more precise decision. Among those strategies, the modification of the normalization process in the multiple-criteria decision-making algorithm is still a question due to the confrontation of many normalization tools. Normalization is the basic action in defining and solving a MADM problem and a MADM model. Normalization is the first, also necessary, step in solving, i.e. the application of a MADM method. It is a fact that the selection of normalization methods has a direct effect on the results. One of the latest normalization methods introduced is the Logarithmic Normalization (LN) method. This new method has a distinguished advantage, reflecting in that a sum of the normalized values of criteria always equals 1. This normalization method had never been applied in any MADM methods before. This research study is focused on the analysis of the classical MADM methods based on logarithmic normalization. VIKOR and TOPSIS, as the two famous MADM methods, were selected for this reanalysis research study. Two numerical examples were checked in both methods, based on both the classical and the novel ways based on the LN. The results indicate that there are differences between the two approaches. Eventually, a sensitivity analysis is also designed to illustrate the reliability of the final results.*

1. INTRODUCTION

In a decision-making problem, there are several elements that influence the preciseness and accuracy of the final solution. In other words, every MADM technique has a different functionality, specification and applicability, which undoubtedly affect the decision-making process, the evaluation system rout and the final priority of alternatives. The point distinguishing between those techniques is the composition of the elements, such as decision variables, normalization tools, their attributes and weights and the computation of the final solution. In the same manner, every optimization problem is recognized as having different structures and elements, incommensurable variables, conflicting development objectives and constraints. Thus, multi-criteria optimization techniques represent an appropriate tool in the ranking or selection of proper alternatives out of a pool of feasible choices in the presence of multiple, and usually conflicting, criteria. As the most significant elements of a typical MADM problem, criteria are naturally stated based on different units of measurement and an optimization direction, such as a meter, a capacity, a litre, a dollar, etc. Moreover, the "benefit" or "non-benefit" orientation of criteria depends on their nature. All MADM techniques are applied in the normalization process [1].

The normalization process is a process of making comparable scales for criteria values, and different methods utilize different approaches to normalization. Multiple attribute decision-making (MADM) frameworks vary from the simple approaches based on a small amount of data to the methods based on mathematical simulation and programming techniques, requiring extensive information for each criterion, and on the decision-maker's preferences as well [2,3,4,5]. A typical decision matrix in a MADM problem contains alternatives and criteria. Criteria have different scales and optimization objectives. In order to avoid difficulties caused by their different dimensions, such criteria values are transformed (or normalized). When the normalization process is completed, it is possible to evaluate criteria by weighting factors. Any MADM model may lack in the delivery of the absolute optimum solution, and normalization norms within the solution methods may fail to reveal the actual decision. In addition to this, different normalization techniques may yield different solutions and, therefore, may cause a deviation from originally recommended solutions [6]. All in all, the problem of how these essential elements (for instance, attributes or criteria) act and how any modifications in the structure and the anatomy will generate more efficient and better outcomes is still a question to answer. Experts and professionals in this field have made attempts to fill this gap and provide reasonable answers through several academic research studies. In this study, a new normalization tool is applied to the two MADM methods called VIKOR and TOPSIS in order to check the consistency of the results.

Many theories have been established so as to express transformation through the normalization procedure. Vector normalization, linear normalization, non-monotonic normalization, Weitendorf's linear normalization (WLN) method, the Jüttler-Körth normalization (JKN) method and the Peldschus non-linear normalization (NLN) method [2,7,8,9] are the normalization tools most applied and employed by scholars. Vafaei et al [10] applied six different normalizations methods to evaluate TOPSIS in all the possible ways. They didn't consider LN in the study. Zadeh Sarraf et al [11] used statistical normalization for evaluating TOPSIS method in a more statistical situation.

MCDM methods are categorized based on their performance. The methods based on a reference point or goal, such as the reference point methods as TOPSIS and VIKOR, the methods based on an initial qualitative assessment as the AHP and the methods based on quantitative measurements using a few attributes to compare alternatives (the comparison preference method) as ELECTRE [12] and PROMETHEE are the three basic classifications of the MADM methods. Apparently, with its special characteristics and definitions, every model contains different normalization methods. For the TOPSIS method, Lai et al [13] proposed vector normalization, only to be followed by a proposal for employing linear normalization in the same multi-criteria method. It is a fact that normalization procedures may affect the final MADM solution. PROMETHEE [14,15] and GRA normally solve a decision problem by using Weitendorf's linear normalization tool, whereas VIKOR [16,17] uses linear normalization to obtain the ranking of alternatives.

In comparison with the other normalization tools, the logarithmic model may be used in the cases when the values of criteria differ considerably from each other and when the solving of problems segregates normalized values more effectively than other methods. A normalized matrix created by the normalization rule yields more stable results in solving multi-criteria decision problems. Furthermore, while in vector normalization (such as TOPSIS & VIKOR) the ratio of the values remains constant in the interval [0,1], the sum of the normalized values of criteria in the logarithmic model is always equal to 1 [7].

In this paper, a novel normalization tool called logarithmic normalization is applied to VIKOR and TOPSIS. The paper intends to apply this normalization to both VIKOR and TOPSIS in order to develop these methods as more powerful methodologies in solving MADM applications. Given the fact that VIKOR's and TOPSIS' final results for the evaluation of alternatives are based on [0,1], this new normalization allows us to absolutely obtain more precise results in the cases of alternatives being higher than 4, the results of the latter method being very near to those of the two former ones. The authors suggest this new framework and extended VIKOR and TOPSIS as the extended models and consider them to be applicable in solving multi-criteria decision-making issues.

## 2. TOPSIS AND VIKOR METHODOLOGIES

As a traditional multi-criteria decision-making method and a MADM problem-ranking tool, the TOPSIS methodology was developed with the aim of reaching non-inferior solutions [18,19]. It has satisfactorily been implemented in different application fields. Its user-friendly anatomy and easy computational algorithm make a decision problem more reliable, thus leading to optimum solutions. In TOPSIS, the best alternative should have the shortest distance ($D^*$) from the positive ideal solution ($v^*$) (which is made up of the best quantity of each criterion regardless of alternatives) and the largest distance ($D^-$) from the negative ideal solution ($v^-$) (which is made up of the worst quantity of each criterion regardless of alternatives). TOPSIS computations begin with an initial pay-off matrix, including criteria and the alternatives accompanying the weights of each criterion. Thereafter, normalized and weighted normalized matrices are detected. The positive ideal solution (PIS) and the negative ideal solution (NIS) should be obtained. A separation measure from PIS and NIS is generated, after which relative closeness to both PIS and NIS is computed. In the final step, the priority ranking order is derived based on such relative closeness to the ideal solutions. The summary account of the steps of the procedure of the TOPSIS method is given in Figure 1.

The VIKOR method was developed in 1998 so as to determine a compromise ranking list of several alternatives with non-commensurable and conflicting criteria and particular weights stability intervals in order to make adjustments to preference stability and the given weights [17,20]. The specific structure of VIKOR introduces it as an applied tool used when decision experts are not able to explain their preferences in the system designing process. In VIKOR, each alternative is measured based on an aggregate function, so the compromise ranking of alternatives is implemented by comparing the measure of closeness to the ideal solution [21]. Any exclusion or inclusion of an alternative could influence the VIKOR ranking results. The VIKOR algorithm prepares the minimum of individual regret and the maximum group utility for the opponent and the majority, respectively [22,23]. In the VIKOR algorithm, $v$ stands for the weight of the strategy of the majority of attributes and is sometimes a useful means to examine the performance of VIKOR. One of VIKOR's characteristics reflects in the fact that the aggregation function is always closest to the best solutions, whereas in TOPSIS, the CC of materials are not necessarily always very close to the ideal values [24,25]. The interactivity of VIKOR allows decision-makers to participate in and control the decision-making process (by means of weights). The aggregating (compound) function is used extremely cautiously since it includes a comparison of potentially incomparable quantities (non-commensurable criteria or indicators). Normalization in VIKOR is performed in order to eliminate the units of indicators; so, all indicators are dimensionless [26]. Every MCDM method will start from a decision matrix including alternatives and criteria, and the mathematical steps of TOPSIS and VIKOR are presented in Figure 1.

TOPSIS has become a key and strategic means and has been launched so as to investigate complex decision-making problems. Several applications acquired TOPSIS as the sole method, the integrated approach, whereas in others, the same was amended for the purpose of obtaining excellent outcomes. In the literature, the reformed type of TOPSIS is presented by practitioners. The extended TOPSIS model appeared in a study [19], determining the weights of decision-makers in a group decision-making environment, in which decision information is expressed by an interval value matrix. The traditional TOPSIS model begins with the normalization process and then defines a weighted normalized decision matrix. In fact, the traditional model believes that the influence of weights should occur in this step in a normalized matrix. However, how do weights exert an influence on a decision matrix and where is the catch? On the one hand, the overall performance of alternatives is estimated by using the Euclidean distance, whereas on the other, a notion is made that weights should affect this distance, for which reason criteria weights should be incorporated in the distance measurement. The extended VIKOR model and, actually, a generalization of VIKOR, are presented by [27,28,29,30]. In order to verify the extended TOPSIS and VIKOR models, a logarithmic normalization tool is applied as follows:

$$f_{ij} = \frac{\ln(x_{ij})}{\ln\left(\prod_{i=1}^{n} x_{ij}\right)} \tag{1}$$

All in all, the previous studies claimed the extended and modified VIKOR and TOPSIS models differently from the weighting process, the aggregation process and the normalization tool can improve the quality and accuracy of a decision. So, the application of a new normalization tool to a MADM problem is examined in this paper in order to observe the results.

## 3. CASE STUDY

In order to verify the proposed approach, the two numerical examples based on a real case are illustrated in this section. The proposed model is implemented in the MHV construction company, located in Madrid, Spain. For approximately 20 years, the company has been participating in the adaptation, designing and construction of buildings, residential apartments and commercial centers projects in Madrid and its urban areas. Now, due to the new rule of the regional government and the ministry of industry, it is compulsory for small firms that they should assure and guarantee the procurement process. Therefore, all further legalization and formal permission of the MHV company is bound by the establishment of the supply evaluation system.

The company should address pieces of evidence in order to approve a safe, secure and environmentally friendly project. Currently, MHV performs its purchasing operations through the official suppliers who are supposed to meet the main ten requirements, such as the foundation, consulting and geometrical measurement, electricity equipment and raw materials, such as wood, cement, iron, etc. for various projects. Among them, a pre-evaluation was carried out, and a decision was made on releasing a list including 4 suppliers (A1 to A4) to be evaluated through a questionnaire.

The questionnaire was designed and distributed among the experts anonymously. The supplier evaluation team consisted of the CEO of the company, two architects (with more than 10 years of experience each) and a purchase assistant. The experts were responsible for making a judgment and offering their score rating for each candidate supplier by applying certain criteria. The evaluation criteria were the offered price, the perceived quality, the experience, the delivery conditions, and the environmental certificates. The primal decision-making matrix was developed as shown in Table 1 after the data had been collected and after the analysis had been made.

**Table 1** The decision matrix

| Alternatives | C1 | C2 | C3 | C4 | C5 |
|---|---|---|---|---|---|
| A1 | 8 | 7 | 7 | 9 | 8 |
| A2 | 7 | 9 | 8 | 7 | 8 |
| A3 | 8 | 8 | 8 | 6 | 9 |
| A4 | 9 | 6 | 7 | 8 | 7 |

Table 2 shows the nature and weight coefficients of each criterion and the computation of VIKOR and the extended model of VIKOR. Based on Figure 1, a mathematical computation of the methods was done. According to the results of the VIKOR and the extended VIKOR methods, the differences between their respective results are possible to see. The new logarithmic normalization [7] presented the obtained results in a more precise manner in the circumstance in which the alternatives in the initial decision-making matrix are of approximately the same values, which will help to rank alternatives more precisely. The computation should be performed for TOPSIS. The procedure and the results are illustrated in Table 3. Unlike the VIKOR ranking scores, the TOPSIS algorithm output for both the classical and the extended models is the same in this example. The brief algorithm of both methodologies is shown in Fig. 1.

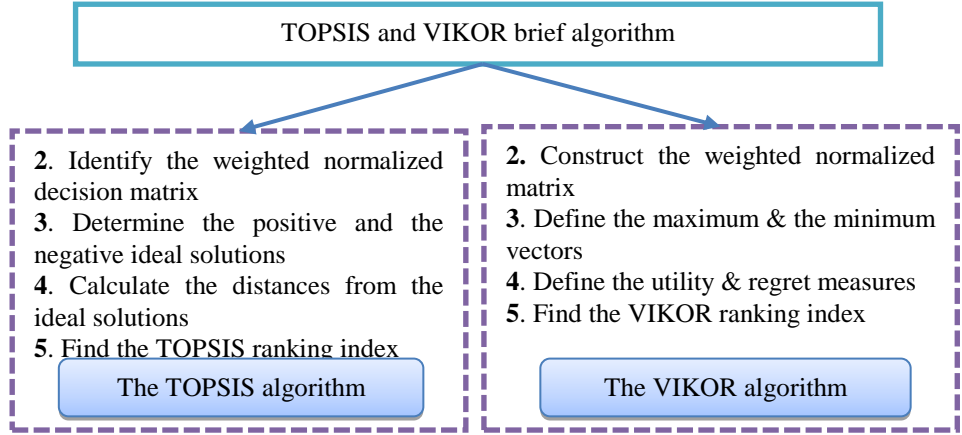

**Fig 1.** The construction of the normalized decision matrix

The decision matrix is the same for both VIKOR and TOPSIS, as shown in Table 1. In this section, both the VIKOR and the TOPSIS calculations are presented in the two mentioned manners, the first related to the classical way to solve the MADM model, and the second being based on Logarithmic Normalization (LN). All the information about the VIKOR calculations is given in Table 2 and that about the TOPSIS calculations in Table 3.

As is illustrated in Tables 2 and 3, the priority of the alternatives has changed, but no changes are perceived for TOPSIS.

In the second example, the different condition of the decision matrix is proposed, implying six criteria and eight alternatives. Table 4 shows the decision table for Example 2. Likewise Example 1, the TOPSIS and the VIKOR methods were adopted as the classical algorithm, including logarithmic normalization as well. The results for each method are presented in Tables 5 and 6, respectively. In this example, we may notice that the TOPSIS ranking changed slightly after the new normalization tool had been applied. However, this never happened in the case of VIKOR, as the same ranking resulted from both the classical and the modified VIKOR algorithms.

**Table 2** The final results of VIKOR

| Crit. | C1 | C2 | C3 | C4 | C5 | C1 | C2 | C3 | C4 | C5 |
|---|---|---|---|---|---|---|---|---|---|---|
|  | Max | Max | Max | Max | Max | Max | Max | Max | Max | Max |
| Weight | 0.197 | 0.163 | 0.176 | 0.197 | 0.267 | 0.197 | 0.163 | 0.176 | 0.197 | 0.267 |
|  | Original VIKOR method | | | | | Modified VIKOR | | | | |
| Alt. | Normalized matrix | | | | | Normalized matrix | | | | |
| A1 | 0.498 | 0.461 | 0.465 | 0.593 | 0.498 | 0.250 | 0.243 | 0.242 | 0.274 | 0.250 |
| A2 | 0.435 | 0.593 | 0.532 | 0.461 | 0.498 | 0.235 | 0.274 | 0.258 | 0.243 | 0.250 |
| A3 | 0.498 | 0.527 | 0.532 | 0.395 | 0.560 | 0.250 | 0.259 | 0.258 | 0.224 | 0.265 |
| A4 | 0.560 | 0.395 | 0.465 | 0.527 | 0.435 | 0.265 | 0.224 | 0.242 | 0.259 | 0.235 |
| f* | 0.560 | 0.593 | 0.532 | 0.593 | 0.560 | 0.265 | 0.274 | 0.258 | 0.274 | 0.265 |
| f- | 0.435 | 0.395 | 0.465 | 0.395 | 0.435 | 0.235 | 0.224 | 0.242 | 0.224 | 0.235 |
|  | Si | Ri | v | Qi | Rank | Si | Ri | v | Qi | Rank |
| A1 | 0.514 | 0.176 | 0.5 | 0.256 | 2 | 0.474 | 0.176 | 0.5 | 0.203 | 3 |
| A2 | 0.462 | 0.197 | 0.5 | 0.274 | 3 | 0.396 | 0.197 | 0.5 | 0.197 | 2 |
| A3 | 0.349 | 0.197 | 0.5 | 0.098 | 1 | 0.344 | 0.197 | 0.5 | 0.116 | 1 |
| A4 | 0.671 | 0.267 | 0.5 | 1 | 4 | 0.654 | 0.267 | 0.5 | 1 | 4 |

**Table 3** The TOPSIS different ranking for both the classical and the extended models

| Criteria | C1 | C2 | C3 | C4 | C5 | C1 | C2 | C3 | C4 | C5 |
|---|---|---|---|---|---|---|---|---|---|---|
|  | Max | Max | Max | Max | Max | Max | Max | Max | Max | Max |
| Weights | 0.197 | 0.163 | 0.176 | 0.197 | 0.267 | 0.197 | 0.163 | 0.176 | 0.197 | 0.267 |
| Alt. | Original TOPSIS method | | | | | Modified TOPSIS | | | | |
|  | Normalized matrix | | | | | Normalized matrix | | | | |
| A1 | 0.041 | 0.039 | 0.04 | 0.06 | 0.041 | 0.25 | 0.243 | 0.242 | 0.274 | 0.25 |
| A2 | 0.036 | 0.050 | 0.045 | 0.047 | 0.041 | 0.234 | 0.274 | 0.258 | 0.243 | 0.25 |
| A3 | 0.041 | 0.044 | 0.045 | 0.04 | 0.046 | 0.25 | 0.259 | 0.258 | 0.224 | 0.265 |
| A4 | 0.046 | 0.033 | 0.04 | 0.054 | 0.036 | 0.265 | 0.224 | 0.242 | 0.259 | 0.234 |
|  | Weighted normalized matrix | | | | | Weighted normalized matrix | | | | |
| A1 | 0.008 | 0.006 | 0.007 | 0.012 | 0.011 | 0.049 | 0.04 | 0.043 | 0.054 | 0.067 |
| A2 | 0.007 | 0.008 | 0.008 | 0.009 | 0.011 | 0.046 | 0.045 | 0.045 | 0.048 | 0.067 |
| A3 | 0.008 | 0.007 | 0.008 | 0.008 | 0.012 | 0.049 | 0.042 | 0.045 | 0.044 | 0.071 |
| A4 | 0.009 | 0.005 | 0.007 | 0.011 | 0.01 | 0.052 | 0.036 | 0.043 | 0.051 | 0.063 |
|  | D+ | D- | CC | Rank |  | D+ | D- | CC | Rank |  |
| A1 | 0.003 | 0.004 | 0.623 | 1 |  | 0.008 | 0.012 | 0.609 | 1 |  |
| A2 | 0.004 | 0.003 | 0.489 | 2 |  | 0.009 | 0.01 | 0.527 | 2 |  |
| A3 | 0.004 | 0.004 | 0.461 | 3 |  | 0.011 | 0.011 | 0.506 | 3 |  |
| A4 | 0.004 | 0.003 | 0.443 | 4 |  | 0.012 | 0.009 | 0.43 | 4 |  |

**Table 4** The initial decision matrix of the second example

| Alternatives | C1 | C2 | C3 | C4 | C5 | C6 |
|---|---|---|---|---|---|---|
| A1 | 4 | 8 | 8 | 7 | 9 | 8 |
| A2 | 6 | 7 | 7 | 8 | 9 | 6 |
| A3 | 7 | 6 | 5 | 8 | 7 | 4 |
| A4 | 6 | 6 | 4 | 6 | 5 | 4 |
| A5 | 9 | 9 | 4 | 6 | 6 | 7 |
| A6 | 7 | 9 | 8 | 8 | 7 | 8 |
| A7 | 8 | 8 | 9 | 8 | 6 | 9 |
| A8 | 9 | 4 | 7 | 5 | 8 | 6 |

**Table 5** The TOPSIS results and ranking for the second example

|  | $C_1$ | $C_2$ | $C_3$ | $C_4$ | $C_5$ | $C_6$ | $C_1$ | $C_2$ | $C_3$ | $C_4$ | $C_5$ | $C_6$ |
|---|---|---|---|---|---|---|---|---|---|---|---|---|
|  | Max | Max | Min | Min | Max | Max | Max | Max | Min | Min | Max | Max |
| Weight | 0.12 | 0.2 | 0.16 | 0.32 | 0.15 | 0.05 | 0.12 | 0.2 | 0.16 | 0.32 | 0.15 | 0.05 |
| Alt. | Original TOPSIS method |  |  |  |  |  | Modified TOPSIS |  |  |  |  |  |
|  | Normalized matrix |  |  |  |  |  | Normalized matrix |  |  |  |  |  |
| $A_1$ | 0.010 | 0.022 | 0.027 | 0.020 | 0.026 | 0.027 | 0.090 | 0.134 | 0.142 | 0.126 | 0.141 | 0.142 |
| $A_2$ | 0.015 | 0.019 | 0.023 | 0.023 | 0.026 | 0.020 | 0.117 | 0.126 | 0.133 | 0.135 | 0.141 | 0.122 |
| $A_3$ | 0.018 | 0.017 | 0.017 | 0.023 | 0.021 | 0.013 | 0.127 | 0.116 | 0.110 | 0.135 | 0.125 | 0.095 |
| $A_4$ | 0.015 | 0.017 | 0.013 | 0.017 | 0.015 | 0.013 | 0.117 | 0.116 | 0.095 | 0.116 | 0.103 | 0.095 |
| $A_5$ | 0.023 | 0.025 | 0.013 | 0.017 | 0.018 | 0.023 | 0.143 | 0.142 | 0.095 | 0.116 | 0.115 | 0.133 |
| $A_6$ | 0.018 | 0.025 | 0.027 | 0.023 | 0.021 | 0.027 | 0.127 | 0.142 | 0.142 | 0.135 | 0.125 | 0.142 |
| $A_7$ | 0.020 | 0.022 | 0.030 | 0.023 | 0.018 | 0.030 | 0.136 | 0.134 | 0.150 | 0.135 | 0.115 | 0.150 |
| $A_8$ | 0.023 | 0.011 | 0.023 | 0.014 | 0.024 | 0.020 | 0.143 | 0.090 | 0.133 | 0.104 | 0.134 | 0.122 |
|  | Weighted normalized matrix |  |  |  |  |  | Weighted normalized matrix |  |  |  |  |  |
| $A_1$ | 0.0012 | 0.0044 | 0.0043 | 0.0063 | 0.0040 | 0.0013 | 0.0108 | 0.0269 | 0.0227 | 0.0403 | 0.0212 | 0.0071 |
| $A_2$ | 0.0018 | 0.0039 | 0.0037 | 0.0073 | 0.0040 | 0.0010 | 0.0140 | 0.0252 | 0.0213 | 0.0431 | 0.0212 | 0.0061 |
| $A_3$ | 0.0021 | 0.0033 | 0.0027 | 0.0073 | 0.0031 | 0.0007 | 0.0152 | 0.0232 | 0.0176 | 0.0431 | 0.0188 | 0.0047 |
| $A_4$ | 0.0018 | 0.0033 | 0.0021 | 0.0054 | 0.0022 | 0.0007 | 0.0140 | 0.0232 | 0.0152 | 0.0371 | 0.0155 | 0.0047 |
| $A_5$ | 0.0027 | 0.0050 | 0.0021 | 0.0054 | 0.0026 | 0.0012 | 0.0172 | 0.0284 | 0.0152 | 0.0371 | 0.0173 | 0.0066 |
| $A_6$ | 0.0021 | 0.0050 | 0.0043 | 0.0073 | 0.0031 | 0.0013 | 0.0152 | 0.0284 | 0.0227 | 0.0431 | 0.0188 | 0.0071 |
| $A_7$ | 0.0024 | 0.0044 | 0.0048 | 0.0073 | 0.0026 | 0.0015 | 0.0163 | 0.0269 | 0.0240 | 0.0431 | 0.0173 | 0.0075 |
| $A_8$ | 0.0027 | 0.0022 | 0.0037 | 0.0045 | 0.0035 | 0.0010 | 0.0172 | 0.0179 | 0.0213 | 0.0333 | 0.0200 | 0.0061 |
|  | D+ | D- | CC | Rank |  |  | D+ | D- | CC | Rank |  |  |
| $A_1$ | 0.0033 | 0.0031 | 0.488 | 4 |  |  | 0.0122 | 0.0113 | 0.481 | 5 |  |  |
| $A_2$ | 0.0035 | 0.0028 | 0.438 | 6 |  |  | 0.0125 | 0.0102 | 0.451 | 7 |  |  |
| $A_3$ | 0.0035 | 0.0028 | 0.437 | 7 |  |  | 0.0121 | 0.01 | 0.451 | 6 |  |  |
| $A_4$ | 0.0029 | 0.0035 | 0.547 | 2 |  |  | 0.0096 | 0.0124 | 0.563 | 2 |  |  |
| $A_5$ | 0.0017 | 0.0046 | 0.735 | 1 |  |  | 0.0055 | 0.0165 | 0.750 | 1 |  |  |
| $A_6$ | 0.0037 | 0.0032 | 0.465 | 5 |  |  | 0.0128 | 0.0122 | 0.488 | 4 |  |  |
| $A_7$ | 0.0041 | 0.0027 | 0.397 | 8 |  |  | 0.0139 | 0.011 | 0.442 | 8 |  |  |
| $A_8$ | 0.0033 | 0.0036 | 0.523 | 3 |  |  | 0.0123 | 0.0129 | 0.511 | 3 |  |  |

**Table 6** The VIKOR results and ranking for the second example

|  | C1 | C2 | C3 | C4 | C5 | C6 | C1 | C2 | C3 | C4 | C5 | C6 |
|---|---|---|---|---|---|---|---|---|---|---|---|---|
|  | Max | Max | Min | Min | Max | Max | Max | Max | Min | Min | Max | Max |
| $w_j$ | 0.12 | 0.2 | 0.16 | 0.32 | 0.15 | 0.05 | 0.12 | 0.2 | 0.16 | 0.32 | 0.15 | 0.05 |
| Alt. | Original VIKOR method |  |  |  |  |  | Modified VIKOR |  |  |  |  |  |
|  | Normalized matrix |  |  |  |  |  | Normalized matrix |  |  |  |  |  |
| A1 | 0.1971 | 0.3871 | 0.4193 | 0.3491 | 0.4386 | 0.4205 | 0.0904 | 0.1344 | 0.1421 | 0.1259 | 0.1412 | 0.1419 |
| A2 | 0.2956 | 0.3388 | 0.3669 | 0.3990 | 0.4386 | 0.3154 | 0.1168 | 0.1258 | 0.1330 | 0.1345 | 0.1412 | 0.1222 |
| A3 | 0.3449 | 0.2904 | 0.2621 | 0.3990 | 0.3412 | 0.2102 | 0.1269 | 0.1158 | 0.1100 | 0.1345 | 0.1251 | 0.0946 |
| A4 | 0.2956 | 0.2904 | 0.2097 | 0.2993 | 0.2437 | 0.2102 | 0.1168 | 0.1158 | 0.0948 | 0.1159 | 0.1034 | 0.0946 |
| A5 | 0.4434 | 0.4355 | 0.2097 | 0.2993 | 0.2924 | 0.3679 | 0.1433 | 0.1420 | 0.0948 | 0.1159 | 0.1152 | 0.1328 |
| A6 | 0.3449 | 0.4355 | 0.4193 | 0.3990 | 0.3412 | 0.4205 | 0.1269 | 0.1420 | 0.1421 | 0.1345 | 0.1251 | 0.1419 |
| A7 | 0.3941 | 0.3871 | 0.4717 | 0.3990 | 0.2924 | 0.4730 | 0.1356 | 0.1344 | 0.1502 | 0.1345 | 0.1152 | 0.1499 |
| A8 | 0.4434 | 0.1936 | 0.3669 | 0.2494 | 0.3899 | 0.3154 | 0.1433 | 0.0896 | 0.1330 | 0.1041 | 0.1337 | 0.1222 |
|  | Si | Ri | v | Qi | Rank |  | Si | Ri | v | Qi | Rank |  |
| A1 | 0.511 | 0.213 | 0.5 | 0.5691 | 4 |  | 0.522 | 0.229 | 0.5 | 0.6286 | 4 |  |
| A2 | 0.598 | 0.320 | 0.5 | 0.9299 | 6 |  | 0.577 | 0.320 | 0.5 | 0.9321 | 6 |  |
| A3 | 0.645 | 0.320 | 0.5 | 0.9862 | 7 |  | 0.615 | 0.320 | 0.5 | 0.9811 | 7 |  |
| A4 | 0.499 | 0.150 | 0.5 | 0.4013 | 3 |  | 0.484 | 0.150 | 0.5 | 0.3776 | 3 |  |
| A5 | 0.239 | 0.113 | 0.5 | 0.0000 | 1 |  | 0.243 | 0.124 | 0.5 | 0.0000 | 1 |  |
| A6 | 0.581 | 0.320 | 0.5 | 0.9095 | 5 |  | 0.565 | 0.320 | 0.5 | 0.9165 | 5 |  |
| A7 | 0.657 | 0.320 | 0.5 | 1.0000 | 8 |  | 0.630 | 0.320 | 0.5 | 1.0000 | 8 |  |
| A8 | 0.364 | 0.200 | 0.5 | 0.3598 | 2 |  | 0.365 | 0.200 | 0.5 | 0.3518 | 2 |  |

As may be observed in both examples, the top alternative (the 1st-ranked) remained the same. Thereafter, except for the VIKOR results in Case 1, the 2nd- and the 3rd-ranked alternatives were the same. In the second example, VIKOR produced a much more stable ranking (the same ranking order). This issue was somehow different, although still acceptable, for TOPSIS, where the first, the second and the third items are similar, whereas the others are replaced in pairs (ranked the 4th and 5th, 6th and 7th). In the cases in which alternatives are so close to each other, this normalization type can be considered as effective. In those cases, prioritizing is more complicated due to the fact that there are a larger number of alternatives whose values are approximately the same in the initial matrix. This research study shows how this normalization may improve the decision-making process and enhance the quality of the results obtained both in VIKOR and TOPSIS by applying logarithmic normalization.

The reliability of a decision-making problem can be judged by a number of alternatives, a number of criteria, the type of the criteria (either qualitative or quantitative, deterministic, interval or fuzzy), and the type of the normalization tool used [31,32,33,34]. Logarithmic normalization for a decision-making matrix yields more stable results in solving decision-making problems [7]. In some cases, solutions are similar; in some other cases, however, very different results are obtained. Given the ranking orders of modified VIKOR and TOPSIS compared with the original methods, the contribution of this study reflects in its providing help to experts in the establishment of different normalization tools for accomplishing optimal solutions and achieving further effective objectives. The normalization rules are a vital part of each MADM method [34]; normalizations and their applications are important for the formation of a new anatomy of MADM tools.

## 4. COMPRATIVE ANALYSIS AND DISCUSSION

The analysis of the stability of the obtained results was carried out in two parts. In the first part, the sensitivity analysis of the LN/original VIKOR and TOPSIS models was performed by changing the weight coefficients of the criteria. The analysis of the influence of the change in the weight coefficients of the criteria was made through 21 scenarios. In the second part, the analysis of the influence of the dynamic decision-making matrices on the change in the rank of the alternatives was performed. A more detailed overview of the sections of the discussion on the results is presented in the next part of the paper.

### 4.1. Changing the Weights of the Criteria

The first part of the discussion is based on the weights sensitivity analysis according to Kahraman [35]. After identifying the *most important criterion* on the basis of the weights estimated in Example 1 and Example 2, the weights sensitivity analysis was performed by varying the weight of the *most important criterion* so as to observe its effect on the ranking performance of the proposed model. According to Kahraman [35], the weight coefficient of the elasticity ($\alpha_s$) of the most important criterion (for Example 1 – C5 and for Example 2 – C4) was assumed to be one. For the other criteria, the coefficients of elasticity ($\alpha_c$) were estimated and shown in Table 6. The elasticity coefficient was used to express the relative compensation of the values of the other weight coefficients in relation to the changes in the weight of the most important criterion.

**Table 6.** The coefficient of elasticity for changing weights

| Criteria | $\alpha_c$ (*Example* 1) | $\alpha_c$ (*Example* 2) |
|---|---|---|
| C1 | 0.269 | 0.176 |
| C2 | 0.222 | 0.294 |
| C3 | 0.240 | 0.235 |
| C4 | 0.269 | 1.000 |
| C5 | 1.000 | 0.221 |
| C6 | - | 0.074 |

Thus, the limit values of C5 and C4 were obtained, being $-0.267 \leq \Delta x \leq 0.733$ (for Example 1) and $-0.320 \leq \Delta x \leq 0.680$ (for Example 2). On the basis of the defined limit values of the change in the weight coefficients of the most important criteria (C5 and C4), the sensitivity analysis scenarios were defined. The intervals $-0.267 \leq \Delta x \leq 0.733$ and $-0.320 \leq \Delta x \leq 0.680$ were divided into 21 sequences, based on which a total of 21 scenarios were formed. For every scenario, the new values of the weight coefficients were formed, so 21 new groups of the weight coefficients were obtained, as is shown in Table 7.

The new values of the weight coefficients were applied to the analyzed models: VIKOR (original), VIKOR (LN), TOPSIS (original) and TOPSIS (LN). The influence of the new values of the weight coefficients on the change in the ranks of the alternatives was analyzed by using Spearman's correlation coefficient (Fig. 2).

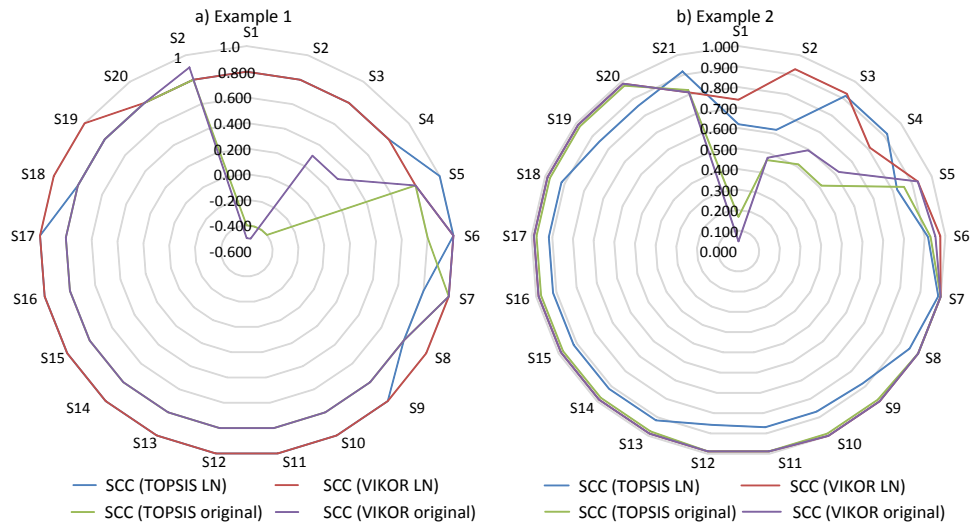

**Fig. 2** The correlation of the ranks through the 21 scenarios – a) Example 1, and b) Example 2

Spearman's correlation coefficient (SCC) was used to determine the statistical significance of the difference between the ranks obtained through the scenarios in the examples 1 and 2. By analyzing the obtained correlation values in the example 1 (Figure 2a), we notice that there is a significantly better correlation of the ranks in the VIKOR LN and TOPSIS LN models in comparison with the original VIKOR and TOPSIS models. These results confirm the fact that the logarithmic model of normalization gives more stable results than vector and linear normalization do [33]. In Scenarios S6-S21, the results of all three normalizations are similar, which is expected, as there are no drastic changes in the C5 criterion in these scenarios.

Similar results were obtained in Example 2 (Fig. 2b). For S1-S5, changing the weight of the C5 criterion resulted in the different ranks of the intermediate alternatives. In the first five scenarios, the logarithmic model of normalization yielded the stable results, with a high correlation of the ranges. This was confirmed by the average value of the correlation coefficient, which was SCC=0.823 for the first five scenarios. Unlike the logarithmic model, the vector and linear models showed a very small value of the SCC, which was 0.508 in the first five scenarios. In the remaining scenarios, i.e. in the scenarios S6-S21, the results of all three normalizations were similar, i.e. all three normalization models showed a high correlation of the ranks. This was confirmed by the average values of Spearman's correlation coefficient: 1) the logarithmic normalization – SCC=0.969; 2) the vector normalization – SCC=0.942; 3) the linear normalization – SCC=0.958. These results confirm the stability of the results obtained by using the logarithmic normalization model.

**4.2. Dynamic Matrices**

When solving a problem, researchers encounter a number of internal and/or external factors, which may change the conditions or limitations of the problem by their actions. Internal changes in a decision matrix, such as the introduction of a new alternative or the elimination of an existing one from a set of the considered alternatives, may lead to a change in the final preferences [36]. Accordingly, in this paper, the performance analysis of the proposed examples (i.e. Examples 1 and 2) was studied in the conditions of a dynamic initial decision matrix. For both examples considered, separate scenarios were formed, and within each scenario, the results of the application of the multi-criteria models were analyzed. For each scenario, a change in the number of the alternatives was made and the ranks obtained were analyzed. The scenarios were formed by removing one inferior (worst) alternative in each scenario from subsequent considerations. Within the scenarios, the remaining alternatives were simultaneously ranked according to the newly matched initial decision matrix.

The initial solution in Example 1 was generated as: (1) Original VIKOR – A3>A1>A2>A4; (1) LN VIKOR – A3>A2>A1>A4 and Original TOPSIS = LNTOPSIS – A1>A2>A3>A4. It is clear that the alternative A4 is the worst option; so, in the first scenario, the alternative A4 was eliminated from the set and a new decision-making matrix was obtained with three alternatives. The new decision matrix was solved again by using MCDM models, and the new rankings of the alternative were obtained (Figure 3). The ranking in the first scenario shows that, for LN TOPSIS and Conventional TOPSIS, the alternative A1 was still the best alternative, and the alternative A3 was the worst. The further implementation of the described procedure resulted in the ranks of the alternatives as shown in Fig. 3.

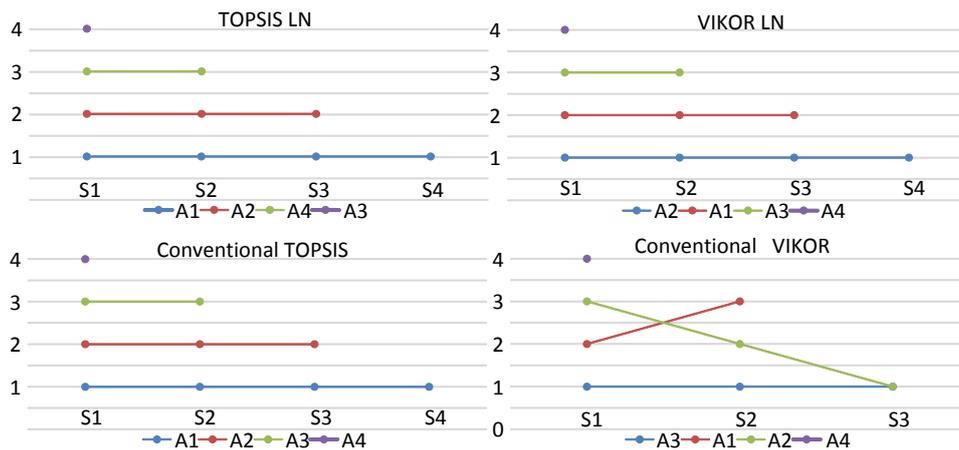

**Fig. 3** The ranks of the alternatives within the dynamic decision matrices – Example 1

Based on the obtained results, it is noted that, when the worst alternative is eliminated, there is no change in the best-ranked alternative in the reorganized matrix for LN TOPSIS, LN VIKOR and Conventional TOPSIS. In Conventional VIKOR, there is a rank reversal problem. After having ranked the third, the alternative A2 became the first-ranked alternative after the second scenario. In the third scenario, the two first-ranked alternatives (namely A3 and A2) appeared. For this reason, the fourth scenario was not implemented. Also, the alternative A1 changed its rank in the scenario S2. In the second scenario, the alternative A1 became third-ranked. This example shows that logarithmic normalization provides stable solutions and is resistant to the rank reversal problem.

A similar analysis was carried out for Example 2 (Figure 4). The analysis was conducted through a total of eight scenarios. In the first scenario, the alternative A7 was eliminated as the worst. The analysis has shown that the LN VIKOR and LN TOPSIS models are stabile in a dynamic environment. In the modifications of the initial matrix, which were made through the elimination of the worst-case option, it was noticed that the LN model did not lead to changes in the ranking (a rank reversal) among the alternatives. The alternative A5 remained the best-ranked across all the scenarios, which confirmed the robustness and accuracy of the ranking alternatives in the dynamic environment. However, in both conventional models (TOPSIS and VIKOR), there is a rank reversal through the scenarios.

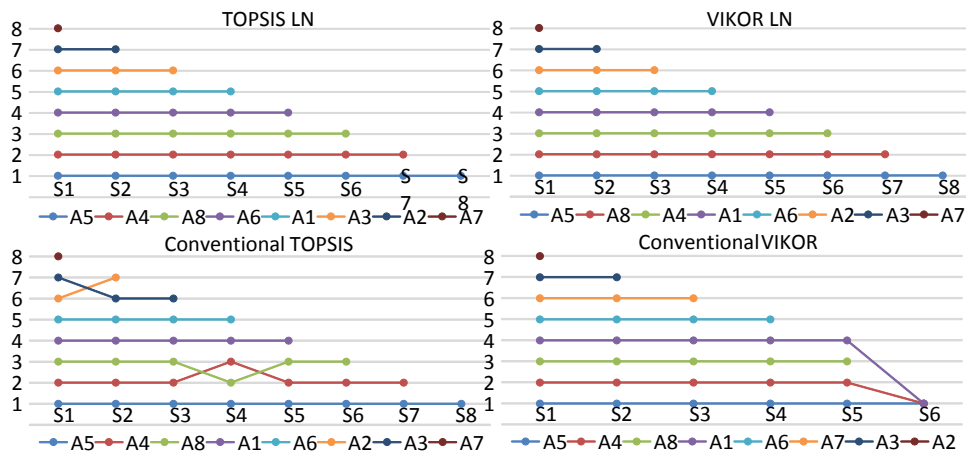

**Fig. 4** The ranks of the alternatives within the dynamic decision matrices – Example 2

These results show the stability of the logarithmic normalization model and its resistance to the rank reversal problem. On these two examples, the logarithmic model of normalization showed significant stability in comparison with the conventional normalization models (the vector and the linear normalization models). In a large number of tests, the ranks obtained by applying LN were equal to the final values (Tables 2, 3, 5 and 6), i.e. there is no rank reversal. This allows us to conclude that logarithmic normalization generates more reliable results than the traditional (vector and linear) normalization models do. Also, it is necessary that the fact that this is a new normalization model which has yet to show its advantages through further empirical research in the field of multi-criteria decision-making should be recognized.

5. Conclusion

The MADM methods, such as TOPSIS, VIKOR and other methods, are still being developed and modified by the addition of other perspectives and approaches. The anatomy of each MADM method can be rebuilt by varying the weight replacement process, the aggregating process, and change in the normalization tool. Normalization allows for a direct comparison of diverse criteria by eliminating the dimensional units that are different. The accuracy of the normalization process is very significant because it must address diverse criteria and objectives in real MADM cases.

As one of the latest normalization methods, the logarithmic normalization method was first presented in 2008. This research study is an attempt to carry out a reanalysis of the two MADM methods based on this new normalization. TOPSIS and VIKOR were selected as the two well-known MADM methods for the purpose of achieving this goal, and the numerical examples were presented in order to illustrate the comparative results. The fact that values are exactly equal to 1 is the advantage of this new normalization method that triggered off the conducting of this research study. The authors believe that, by applying LN, results can be different in the cases in which alternatives are close to one another.

In the first example, the ranking results obtained by applying the VIKOR method changed; for the ranking results obtained by applying the TOPSIS method, however, they were still the same, whereas the rank reversal problem occurred in the second example. LN seems to have had a positive effect on both VIKOR and TOPSIS. Technically, this normalization tool can achieve considerable outcomes, comparing to other normalization approaches. Using both VIKOR and TOPSIS on other examples and other cases in future research studies would be highly appreciated. This research study shows that the accuracy of the normalization methods is effective and sensitive to results and methodologies. Hence, although normalization methods generate the values that are but slightly different from each other, those small differences may have significant consequences for the quality of decision-making and the final decision when selecting among feasible alternatives. This new normalization tool can be applied in other classical MADM methods, such as COPRAS, WASPAS, ARAS, etc. and new methods will also be introduced in the future.

References

1. Jassbi, J.J., Ribeiro, R.A., Varela, L.R., 2014, *Dynamic MCDM with future knowledge for supplier selection*, Journal of Decision Systems, 23(3), pp. 232-248.
2. Zavadskas, E.K., Zakarevicius, A., Antucheviciene, J., 2006, *Evaluation of ranking accuracy in multi-criteria decisions*, Informatica, 17(4), pp. 601-618.
3. Chen, T., 2011, *Using hybrid MCDM model for enhancing the participation of teacher in recreational sports*, Journal of Decision Systems, 20(1), pp. 33-49.
4. Köksalan, M., Wallenius, J., Zionts, S., 2013, *An early history of multiple criteria decision making*. Journal of Multi-Criteria Decision Analysis, 20(1-2), pp. 87-94.
5. Ignatius, J., Rahman, A., Yazdani, M., Šaparauskas, J., Haron, S.H., 2016, *An integrated fuzzy ANP–QFD approach for green building assessment*, Journal of Civil Engineering and Management, 22(4), pp. 551-563.

6. Chatterjee, P., Chakraborty, S., 2014, *Investigating the effect of normalization norms in flexible manufacturing system selection using multi criteria decision making methods*, Journal of engineering science and technology review, 7(3), pp. 141-150.
7. Zavadskas, E. K., Turskis, Z., 2008, *A new logarithmic normalization method in games theory*, Informatica, 19(2), pp. 303-314.
8. Eftekhary, M., Gholami, P., Safari, S., Shojaee, M., 2012, *ranking normalization methods for improving the accuracy of SVM algorithm by DEA method*, Modern applied science, 6(10), pp. 26-36.
9. Peldschus, F., Vaigauskas, E., Zavadskas, E.K., 1983, Technologische entscheidungen bei der berücksichtigung mehrerer Ziehle.
10. Vafaei, N., Ribeiro, R.A., Camarinha-Matos, L. M., 2018, *Data normalisation techniques in decision making: case study with TOPSIS method*, International journal of information and decision sciences, 10(1), pp. 19-38.
11. Zadeh Sarraf, A., Mohaghar, A., Bazargani, H., 2013, *Developing TOPSIS method using statistical normalization for selecting Knowledge management strategies*, Journal of industrial engineering and management, 6(4), pp. 860-875.
12. Álvarez Carrillo, P.A., Leyva López, J.C., Sánchez Castañeda, M.D.L.D., 2015, *An empirical study of the consequences of coordination modes on supporting multicriteria group decision aid methodologies*, Journal of decision systems, 24(4), pp. 383-405.
13. Lai, Y.J., Liu, T.Y., Hwang, C.L., 1994, TOPSIS for MODM, *European journal of operational research*, 76(3), pp. 486-500.
14. Mareschal, B., Mertens, D., 1992, *Banks a multicriteria, PROMETHEE-based, decision support system for the evaluation of the international banking sector*, Journal of decision systems, 1(2-3), pp. 175-189.
15. Behzadian, M., Kazemzadeh, R.B., Albadvi, A., Aghdasi, M., 2010, *PROMETHEE: A comprehensive literature review on methodologies and applications*, European journal of operational research, 200(1), pp. 198-215.
16. Opricovic, S., Tzeng, G.H., 2003, *Fuzzy multi-criteria model for post-earthquake land-use planning*, Natural hazards review, 4(2), 59-64.
17. Opricovic, S., Tzeng, G.H., 2004, *Compromise solution by MCDM methods: A comparative analysis of VIKOR and TOPSIS*, European journal of operational research, 156(2), pp. 445-455.
18. Wang, P., Zhu, Z., Wang, Y., 2016, *A novel hybrid MCDM model combining the SAW, TOPSIS and GRA methods based on experimental design*, Information Sciences, 345, pp. 27-45.
19. Omar, M.N., Fayek, A.R., 2016, *A TOPSIS-based approach for prioritized aggregation in multi-criteria decision-making problems*, Journal of multi-criteria decision analysis, doi: 10.1002/mcda.1561.
20. Opricovic, S., 1998, *Multi-criteria optimization of civil engineering systems*, Faculty of Civil Engineering, Belgrade, 2(1), pp. 5-21.
21. Yazdani, M., Graeml, F.R., 2014, *VIKOR and its applications: A state-of-the-art survey*, International Journal of Strategic Decision Sciences, 5(2), 56-83.
22. Noureddine, M., Ristic, M., 2019, *Route planning for hazardous materials transportation: Multicriteria decision making approach*, Decision making: applications in management and engineering, 2(1), pp. 66-85.
23. Bozanic, D., Tešić, D., Kočić, J., 2019, *Multi-criteria FUCOM – Fuzzy MABAC model for the selection of location for construction of single-span bailey bridge*, Decision making: applications in management and engineering, 2(1), pp. 132-146.
24. Nunić, Z., 2018, *Evaluation and selection of Manufacturer PVC carpentry using FUCOM-MABAC model*, Operational research in engineering sciences: Theory and applications, 1(1), pp. 13-28.

25. Erceg, Ž., Mularifović, F., 2019, *Integrated MCDM model for processes optimization in supply chain management in wood company*, Operational research in engineering sciences: Theory and applications, 2(1), pp. 37-50.
26. Opricovic, S., 2009, *A compromise solution in water resources planning*, Water resources management, 23(8), pp. 1549-1561.
27. Opricovic, S., Tzeng, G.H., 2007, *Extended VIKOR method in comparison with outranking methods*, European journal of operational research, 178(2), pp. 514-529.
28. Pamucar, D., Bozanic, D., Lukovac, V., Komazec, N., 2018, *Normalized weighted geometric bonferroni mean operator of interval rough numbers – application in interval rough DEMATEL-COPRAS*, Facta Universitatis-Series Mechanical Engineering, 16(2), pp. 171-191.
29. Pamučar, D., Božanić, D., 2019, *Selection of a location for the development of multimodal logistics center: Application of single-valued neutrosophic MABAC model*, Operational research in engineering sciences: Theory and applications, 2(2), pp. 55-71.
30. Sayadi, M.K., Heydari, M., Shahanaghi, K., 2009, *Extension of VIKOR method for decision making problem with interval numbers*, Applied Mathematical Modelling, 33(5), pp. 2257-2262.
31. Jahan, A., Edwards, K.L., 2015, *A state-of-the-art survey on the influence of normalization techniques in ranking: Improving the materials selection process in engineering design*, Materials & design, 65, pp. 335-342.
32. Vafaei, N., Ribeiro, R.A., Camarinha-Matos, L.M., 2016, *Normalization techniques for multi-criteria decision making: Analytical hierarchy process case study*, In Technological innovation for cyber-physical systems, Springer international publishing, pp. 261-269.
33. Mukhametzyanov, I., Pamucar, D., 2018, *A sensitivity analysis in MCDM problems: A statistical approach*, Decision making: applications in management and engineering, 1(2), pp. 51-80.
34. Pamučar, D., Božanić, D., Ranđelović, A., 2017, *Multi-criteria decision making: An example of sensitivity analysis*, *Serbian journal of management*, 11(1), pp. 1-27.
35. Kahraman, Y.R., 2002, R*obust sensitivity analysis for multi-attribute deterministic hierarchical value models*, Storming Media, Ohio.
36. Jankovic, A., Popovic, M., 2019, *Methods for assigning weights to decision makers in group AHP decision-making*, Decision making: applications in management and engineering, 2(1), pp. 147-165.